\documentclass[11pt]{article}

\usepackage[preprint]{acl}

\usepackage{times}
\usepackage{latexsym}

\usepackage[T1]{fontenc}

\usepackage[utf8]{inputenc}

\usepackage{microtype}

\usepackage{inconsolata}

\usepackage{graphicx}

\usepackage{amsmath}
\usepackage{amssymb}
\usepackage{multirow}
\usepackage{xcolor}
\usepackage{array}
\usepackage{booktabs}
\usepackage{comment}
\usepackage{algorithm}
\usepackage{algpseudocode}
\usepackage{tcolorbox}
\usepackage[table]{xcolor}

\definecolor{oursrow}{RGB}{234,242,255}

\title{AlphaMemo: Structured Search-Process Memory for \\ Self-Evolving Alpha Mining Agents}

\author{Hang Yu \\
  University of Sydney \\
  \And
  Zifan Zheng \\
  University of Sydney \\
  \And
  Chun Gan\\
  JD.com, Inc. \\
  \And
  Jeff Z. Pan\\
  University of Edinburgh \\
  \AND
  Tongliang Liu\\
  University of Sydney \\
  \And
  Zhiyong Wang\\
  University of Sydney \\
  \And
  Fengxiang He \\
  University of Edinburgh \\
  }

\begin{document}
\maketitle
\begin{abstract}
LLM agents are promising for alpha mining via combining financial priors, symbolic reasoning, executable factor generation, and feedback-driven refinement. Yet, they face a combinatorial search space, noisy non-stationary feedback, redundant discoveries, and overfitting risks from naively reusing past successes. To address these challenges, we propose AlphaMemo, a self-evolving alpha mining agent with Structured Search-Process Memory. Rather than memorizing only final factors or full trajectories, AlphaMemo records reusable evidence about which edit motifs work or fail under specific parent-factor contexts. It extracts motifs from Abstract Syntax Tree (AST) differences, applies confidence-gated residual memory on top of a search-ledger prior, and uses asymmetric veto control to suppress high-confidence failure patterns. Experiments on CSI 500 and S\&P 500 show improved out-of-sample performance and fixed-budget discovery efficiency, with ablations validating the roles of residual learning, confidence gating, AST-diff motifs, and veto memory. Code is at \url{https://github.com/jarrettyu/sspm}. 
\end{abstract}

\section{Introduction}
\label{sec:intro}

Alpha factor mining seeks predictive and interpretable signals for future asset returns, typically by representing each formulaic alpha as a symbolic expression over price-volume variables and selecting it into a factor pool through backtesting under quality and diversity constraints. The field has evolved from human-designed anomalies and empirical asset-pricing factors \citep{fama1993common,kakushadze2016alphas,gu2020mlassetpricing} to machine-learning predictors and automated symbolic search \citep{chen2021gp,yu2023alphagen,shi2025alphaforge,zhu2025alphaqcm}. This executable-feedback loop makes alpha mining a natural setting for LLM agents, which can use financial priors, translate hypotheses into formulas, repair invalid expressions, and refine candidates from numerical feedback \citep{wang2025alpha,kou2024automate,li2024FAMA,tang2025alphaagent}.
More broadly, self-evolving agents improve by persisting experience across trials \citep{gao2025surveyselfevolving,fang2025comprehensivesurvey}, alpha-mining agents should therefore accumulate search experience rather than restart each generation from scratch.

Reusing search experience requires more than remembering successful formulas. \emph{A high-scoring child does not necessarily imply a useful edit.} It may simply retain a strong parent's signal. Rewarding the edit with the child's full score can therefore reinforce changes that merely preserve a good starting point. Noisy return correlations make individual improvements uncertain, and a predictive child may still be too redundant to enter the factor pool. We seek compact experience about which edits deserve reuse without retrieving full mining trajectories. This is a search-level credit-assignment problem in which the agent must estimate an edit's value relative to comparable parent contexts before reusing it.

To address these challenges, we propose {AlphaMemo}, a self-evolving alpha-mining agent built around \emph{Structured Search-Process Memory} (SSPM).
AlphaMemo keeps the factor library as a search ledger for parent selection and learns reusable edit experience through SSPM.
For each distinct valid child, it adjusts predictive quality using the historical outcomes of children from similarly scored parents. It assigns this credit to the realized structural edit under the parent category, rather than to the requested mutation label.
Confidence control discounts sparse or variable credit estimates, while admission-failure memory penalizes edits that repeatedly violate quality or redundancy constraints.
Together, these estimates guide which edits the agent should reuse under each parent context.
The task-specific design ties memory to executable price-volume formulas, inherited parent signals, temporal predictive feedback, and constrained factor admission.

\begin{figure*}[t!]
  \centering
  \includegraphics[width=\textwidth]{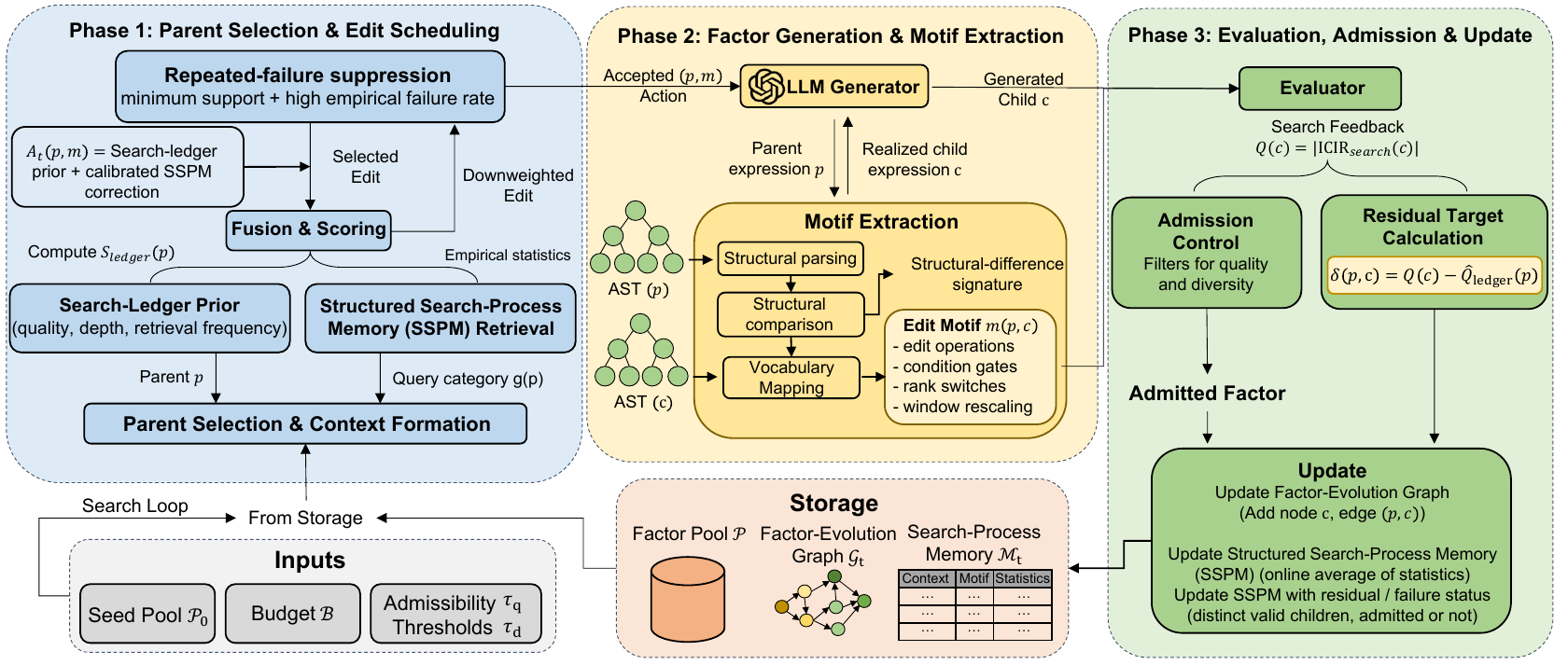}
  \caption{Architecture of AlphaMemo. The search ledger selects parent factors. SSPM attributes feedback to realized structural edits, estimates parent-adjusted credit, and guides subsequent edit proposals using confidence and admission-failure evidence.}  \label{fig:graph}
\end{figure*}

To represent what the LLM actually changed, AlphaMemo derives a lightweight structural motif from changes in operators, features, temporal windows, and expression nesting between a parent and its realized child.
The motif provides a reusable unit for assigning credit across related formula edits.
AlphaMemo also applies an asymmetric process veto. Positive residual evidence remains a soft preference, whereas a context-edit pair is strongly suppressed only after repeated failures exceed a conservative empirical threshold.
This asymmetry reduces repeated expenditure on observed high-failure actions without treating every negative observation as a permanent rule.

Our main contributions are as follows:
\begin{itemize}
    \setlength{\itemsep}{0pt}
    \setlength{\parskip}{0pt}
    \setlength{\parsep}{0pt}
    \item We propose AlphaMemo, a self-evolving alpha-mining framework that addresses search-level credit assignment through SSPM by reusing edit-level experience.
    \item We design SSPM as context-conditioned agentic memory that records realized structural edit motifs, aggregates reference-adjusted outcomes, and controls experience reuse through confidence gates and admission-failure suppression.
    \item Extensive experiments on CSI 500 and S\&P 500 examine predictive and portfolio utility, including exploratory market-specific settings. The CSI 500 ablations favor parent-adjusted credit, while sensitivity results highlight calibrating when and how strongly experience guides search.
\end{itemize}

\section{Related Work}
\label{sec:related}

\paragraph{Self-evolving agents.}
Self-evolving agents update prompts, tools, memories, policies, or model parameters from interaction feedback \citep{gao2025surveyselfevolving,fang2025comprehensivesurvey}. Among these targets, memory is appealing because it enables adaptation without parameter updates. Prior agents store reflections \citep{shinn2023reflexion}, use persistent external state for long-horizon interaction \citep{yao2023react,park2023generative,packer2023memgpt,wang2023voyager}, maintain experience buffers or learned memory operations \citep{liu2025cer,wei2025evo,yan2025memoryr1,yu2026agemem}, and preserve structured intermediate states in tree search, program search, prompt evolution, or code evolution \citep{zhou2023lats,antoniades2025swe,zheng2025mctsahd,guo2024connecting,romera2024mathematical,lee2025evolving,novikov2025alphaevolve}. These methods show that external state can improve test-time behavior, but also raise the local credit-assignment problem of identifying which decision caused success or failure in a multi-step trajectory. \citep{zeng2025turnlevel}. 

AlphaMemo studies this issue in symbolic financial search by storing structured edge-level evidence about which realized edit motifs yield residual gains beyond the base search prior under a given factor-category context. This memory unit links parent-adjusted outcomes, realized structural edits, and evidence-controlled reuse within executable formula evolution.

\paragraph{LLM agents for alpha mining.}
Formulaic alpha mining has moved from genetic programming and reinforcement-learning search over symbolic expressions \citep{zhang2020autoalpha,chen2021gp,yu2023alphagen,shi2025alphaforge,zhu2025alphaqcm,jiang2025alpha} to graph-based, generative, and diversity-aware discovery \citep{li2025alphagat,zhao2025quantfactor,chen2025alphasage,ding2025alphaeval}. LLM-based systems further translate natural-language ideas into executable factors, refine them through feedback, combine generation with evaluation and portfolio construction, or use neural-symbolic and multi-agent workflows \citep{wang2025alpha,kou2024automate,li2024FAMA,tang2025alphaagent,li2025rdagentquant}. Related refiners use feedback, chain-style reasoning, multi-agent search, or Monte Carlo tree search for local formula improvement \citep{luo2025efs,shi2025navigating}. Recent self-evolving alpha-mining systems adapt at different levels, including policy learning \citep{tang2026alphaagentevo}, full hypothesis-expression-code trajectories \citep{han2026quantaalpha}, and experience memories of successful and forbidden patterns \citep{wang2026factorminer}. 

AlphaMemo is complementary to those approaches. It keeps the interpretable formulaic setting and structured factor-library state, but makes self-evolution an inspectable process memory over factor contexts and realized structural edit motifs rather than a policy update, terminal-result memory, or full trajectory store.

\section{Problem Formulation}
\label{sec:formulation}

\paragraph{Alpha mining.}
Let $\mathcal{X}_{\leq t}=\{x_{i,u}:i\in\{1,\ldots,N\},u\leq t\}$ be the available market-feature history for $N$ assets.
A formulaic alpha $f\in\mathcal{F}$ is an executable expression in the supported price--volume operator grammar. It produces scores without observing future prices:
\begin{align}
    (f_{1,t},\ldots,f_{N,t}) &= f(\mathcal{X}_{\leq t}), \\
    r^{(h)}_{i,t} &= \frac{P_{i,t+h+1}}{P_{i,t+1}}-1,
\end{align}
where $P_{i,t}$ is the close price and $h$ is the holding horizon, aligned from the next close. The goal is a predictive, non-redundant factor pool, not an isolated best expression.

\paragraph{Search feedback.}
We write $Q(f)=|\mathrm{ICIR}_{\mathrm{search}}(f)|$ for predictive quality and $\rho_{\mathrm{search}}(f,f')$ for factor correlation on the search period.
A distinct valid child $c$ is admitted to pool $\mathcal{P}_t$ when
\begin{equation}
    Q(c)\geq\tau_q,\quad
    \max_{f\in\mathcal{P}_t}|\rho_{\mathrm{search}}(c,f)|\leq\tau_d,
\end{equation}
with the correlation term zero for an empty pool. Only search-period feedback updates online memory. Downstream predictive and portfolio metrics are defined in Section~\ref{sec:experiments} and Appendix~\ref{app:metrics}.

\paragraph{Factor-library search ledger.}
We model iterative alpha mining as a structured search ledger $\mathcal{G}_t=(\mathcal{V}_t,\mathcal{E}_t)$.
Each node $v\in\mathcal{V}_t$ is an admitted factor, and each directed edge $(p,c)\in\mathcal{E}_t$ records an admitted child generated by editing parent $p$.
A base scorer assigns a parent score $S_{\mathrm{ledger}}(p)$ based on factor quality, depth, and retrieval frequency. The admission rule separately enforces diversity.
This graph provides the selectable library-state prior. A separate event record preserves every evaluated attempt for auditing. Credit estimation uses distinct valid children, including valid children rejected by admission.

\paragraph{Structured search-process memory.}
AlphaMemo assigns search-level credit to a realized parent--child transition. The question is which local edits deserve reuse after accounting for the quality inherited from the parent.
For a distinct valid child $c$ generated from parent $p$, we extract:
\begin{equation}
    e = (g(p), m(p,c), \delta(p,c), s(p,c)),
\end{equation}
where $g(p)$ is the parent category, $m(p,c)$ is the structural edit motif derived from the parent-child expression difference, $\delta(p,c)$ is the reference-adjusted outcome, and $s(p,c)$ records whether the child passes admission.
The memory table aggregates these edge-level observations so that later search can ask which edit motifs have historically produced reliable residual gains in the current factor category.

Both retrieval and updates use the same parent context $g(p)$, even if the child changes category. Parent quality is separately bucketed as $b_q(p)\in\{\text{low},\text{medium},\text{high}\}$ to estimate the credit reference. Invalid and syntactically duplicate expressions do not update numerical credit or failure counts. This distinguishes learning from executable edits from tracking generator errors.

\paragraph{Residual.}
The residual signal is defined as
\begin{equation}
\label{eq:residual_target}
    \delta(p,c) = Q(c) - \widehat{Q}_{\mathrm{ledger}}(p),
\end{equation}
where $\widehat{Q}_{\mathrm{ledger}}(p)$ is the empirical child-quality baseline for the quality bucket of parent $p$.
Let $\mathcal{H}_t(b_q(p))$ denote previously evaluated, distinct valid children whose parents fall in the same bucket. Before incorporating the current child, the baseline is
\begin{equation}
\label{eq:ledger_baseline}
    \widehat{Q}_{\mathrm{ledger}}(p)
    =
    \frac{1}{|\mathcal{H}_t(b_q(p))|}
    \sum_{c_i\in\mathcal{H}_t(b_q(p))}Q(c_i),
\end{equation}
with $\widehat{Q}_{\mathrm{ledger}}(p)=Q(p)$ when the bucket has no prior observation.
This empirical reference reduces the tendency to credit an edit merely because it was applied to a strong parent.
It estimates relative editing experience but does not identify causal effect, as coarse buckets and noisy financial feedback can leave residual confounding.

\section{AlphaMemo}
\label{sec:method}

AlphaMemo organizes self-evolution around search-level credit assignment. It identifies the realized edit and estimates its value relative to its parent context, and then decides when that experience is reliable enough to reuse.
The search ledger selects parents, while SSPM guides which edits to propose for each selected parent.
At iteration $t$, the agent maintains a selectable factor ledger $\mathcal{G}_t$, an admitted factor pool $\mathcal{P}_t$, and a process memory $\mathcal{M}_t$.
It generates and evaluates batches of children, records every attempt, and updates credit from distinct valid transitions. Only admitted children enter the selectable ledger and factor pool.
Figure~\ref{fig:algorithm_chart} gives the per-iteration decision flow, and Algorithm~\ref{alg:alphamemo_appendix} in Appendix~\ref{app:algorithm_theory} provides the corresponding pseudocode and derivation details.

\begin{figure}[h]
\centering
\includegraphics[width=\linewidth]{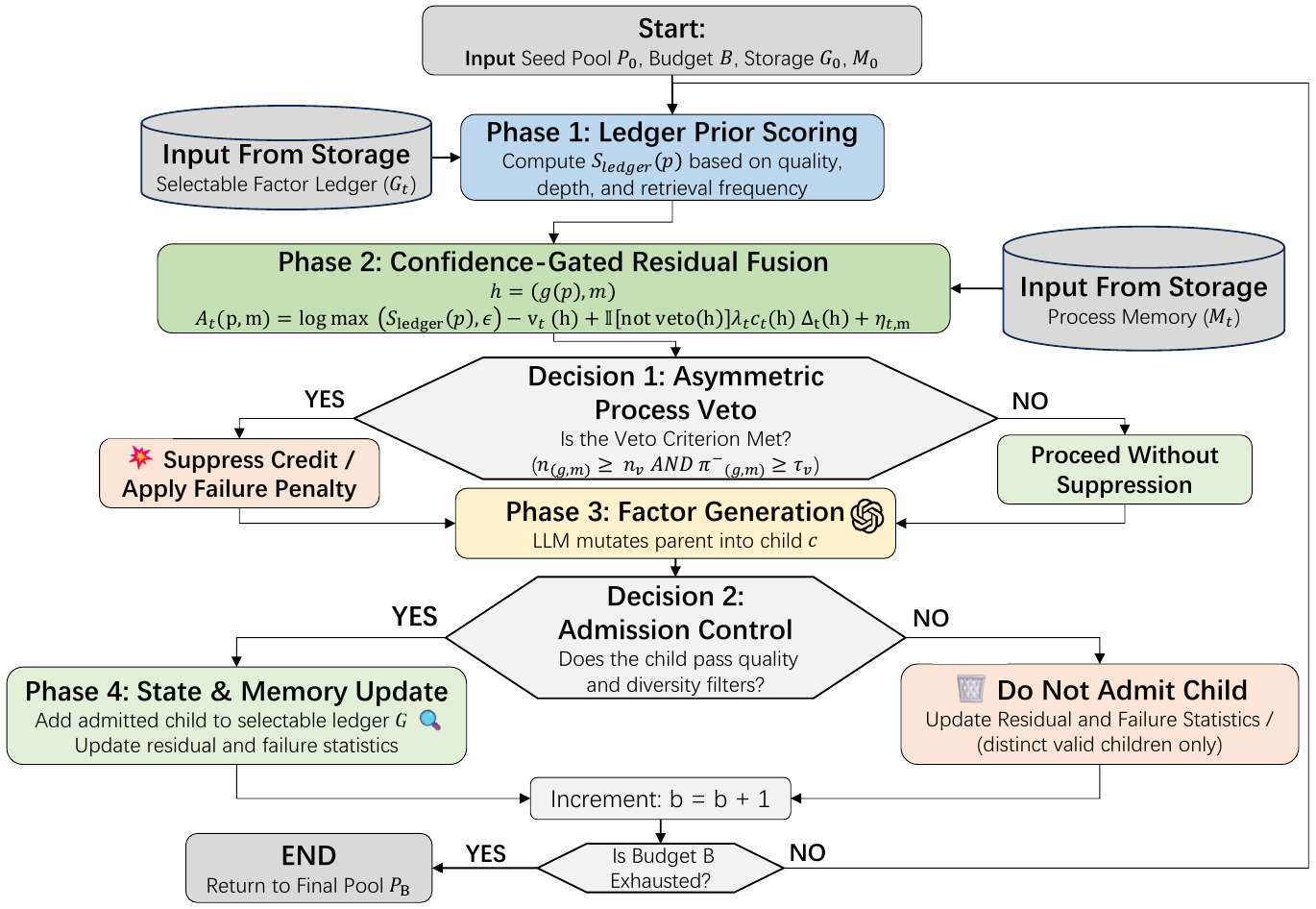}
\caption{Per-iteration decision flow of AlphaMemo. The search ledger retrieves parents, SSPM guides their edit proposals, and executable evaluation provides feedback for admission and credit updates.}
\label{fig:algorithm_chart}
\end{figure}

\subsection{Parent-Edit Action Space}

The action space is defined over pairs $(p,m)$, where $p$ is a parent factor in the search ledger and $m$ is an edit motif.
This differs from a standard parent-only retriever.
The parent determines where to search in the current library state, while the motif determines how to modify the parent.
Examples of high-level motifs include adding a conditional gate, rescaling a temporal window, replacing a ranking operator, adding an interaction term, smoothing a noisy signal, or reversing a sign.
In the reported configuration, we use ten motif labels: nine named edit types plus an ``other'' bucket for rare or ambiguous edits.

Parents are selected by the base score $S_{\mathrm{ledger}}(p)$ before memory-guided edit selection.
For each selected parent, we sample a small set of motifs and retrieve estimates associated with $(g(p),m)$.
For compactness, let $h_{p,m}=(g(p),m)$ denote this memory key.
The recorded action score is:
\begin{align}
\label{eq:cgrm_score}
    A_t(p,m)
    ={}& \log\max(S_{\mathrm{ledger}}(p),\epsilon)
    -v_t(h_{p,m}) \notag \\
    &+\mathbb{I}[\neg\mathrm{Veto}(h_{p,m})]
    \lambda_t \notag \\
    &\qquad \cdot c_t(h_{p,m})\Delta_t(h_{p,m})+\eta_{t,m}.
\end{align}
Here $\Delta_t(g,m)$ is mean residual credit, $c_t(g,m)\in[0,1]$ is confidence, $v_t(g,m)$ is a failure penalty, and $\eta_{t,m}\sim\mathcal{N}(0,0.01^2)$ maintains stochastic exploration.
Within a selected parent, the ledger term is constant, so only the memory terms and proposal noise rank sampled motifs. This score does not globally rerank parents.

\subsection{Structured Search-Process Memory (SSPM)}

SSPM is the confidence-gated residual memory that records how local edit motifs behave under coarse factor-category contexts.
For each category-motif pair $(g,m)$, it stores compact empirical statistics:
\begin{equation}
    \mathcal{M}_t[g,m] =
    \{n_{g,m}, \mu_{g,m}, \sigma^2_{g,m}, n^+_{g,m}, n^-_{g,m}\},
\end{equation}
where $n$ counts distinct valid transitions, $\mu$ and $\sigma^2$ summarize residual outcomes, and $n^+$ and $n^-$ count admitted and rejected children.
After evaluating a distinct valid child $c$, we compute Equation~\ref{eq:residual_target} and update the entry keyed by $(g(p),m(p,c))$.

The confidence gate prevents early noisy observations from affecting search and downweights unstable residual estimates.
The reported implementation uses a count, outcome-consistency, and variance gate. Let $\hat p=(n^++1)/(n+2)$ and let $H_2$ be binary entropy. For $n\geq2$,
\begin{equation}
\label{eq:confidence}
\begin{split}
    c_t(g,m)={}&\min\left(1,\frac{n_{g,m}}{n_{\mathrm{conf}}}\right)
    \left(1-\frac{H_2(\hat p_{g,m})}{2}\right)\\
    &\cdot\max\left(0,1-\frac{\sigma_{g,m}}
    {|\mu_{g,m}|+0.03+\epsilon}\right),
\end{split}
\end{equation}
and $c_t(g,m)=0$ when $n<2$. During warmup, motifs are sampled uniformly. Thereafter, random exploration is retained with probability 0.35. Other proposals use four sampled motifs and a fixed memory weight. Parent selection continues to use the ledger throughout.

\subsection{Realized Structural Edit Motifs}

LLM-generated edits are often not well captured by manually designed mutation labels.
For example, a model may simultaneously add a volume gate, rescale a rolling window, and change a normalization operator.
After evaluator-side parsing, AlphaMemo compares function counts, feature identities, temporal-window constants, other constants and arithmetic operations between parent and child abstract syntax tree profiles. The motif $m(p,c)$ maps this signature to a compact vocabulary:
\begin{align}
    m(p,c) = \phi\big(d_{\mathrm{struct}}(p,c)\big).
\end{align}
The current vocabulary covers condition gates, rank switches, interaction additions, window rescaling, operator substitutions, feature swaps, nesting increases, temporal shifts, normalization changes, and other edits.
The same vocabulary guides generation. A fixed precedence rule assigns one realized motif to a multi-part change without assuming that the requested edit was followed exactly. This coarse profile is not a minimal tree-edit script.
This representation provides two advantages.
First, it assigns credit to the actual symbolic change that the LLM produced.
Second, it can transfer across surface-level expression variants that share the same edit behavior.

\subsection{Asymmetric Process Veto}

AlphaMemo uses failure evidence conservatively rather than treating every negative observation as a reusable rule.
Positive residual memories only enter as a soft additive term in Equation~\ref{eq:cgrm_score}.
Repeatedly unsuccessful category-motif pairs receive a stronger penalty before generation.
Let $\pi^-_{g,m}=n^-_{g,m}/n_{g,m}$ be the empirical admission-failure rate among distinct valid children. Before five observations the failure severity is zero. Thereafter it is $\pi^-_{g,m}$ for a flagged cell and $\pi^-_{g,m}/2$ otherwise. The penalty $v_t(h)$ scales this severity by a failure-memory weight in $(0,1]$.
The veto rule $\mathrm{Veto}(g,m)$ equals
\begin{equation}
\label{eq:veto}
    \mathbb{I}\left[n_{g,m}\geq n_v
    \ \wedge\
    \pi^-_{g,m}\geq\tau_v\right].
\end{equation}
We use $n_v=5$ and $\tau_v=0.8$. A flag suppresses the residual term and applies a finite failure penalty. Flagged motifs remain selectable when alternatives are worse, and random exploration remains available.
This asymmetry is specific to constrained factor admission because an executable edit can repeatedly fail quality or redundancy checks even when its formula is valid.

\subsection{Search Loop}

Each iteration consists of the following steps:
\begin{enumerate}
    \setlength{\itemsep}{0pt}
    \setlength{\parskip}{0pt}
    \setlength{\parsep}{0pt}
    \item Retrieve parents from the ledger and choose motifs by warmup, random exploration, or Equation~\ref{eq:cgrm_score}.
    \item Prompt the LLM with each parent's expression, category, motif, and operator constraints.
    \item Evaluate proposals on the search period and log every attempt. Invalid or duplicate children do not update memory statistics.
    \item For each distinct valid child, update parent-context memory with its realized motif, residual credit, and admission outcome.
    \item Add admitted children and their edges to the ledger for subsequent search.
\end{enumerate}
The ledger records visited factors, while process memory accumulates edit experience for subsequent proposals.

\section{Mechanistic Analysis}
\label{sec:analysis}

The analysis explains how the memory design addresses edit-level credit assignment and what its score control can guarantee.

\subsection{Residual Memory as Search-Prior Correction}

A naive memory prior can be harmful when paired with a strong search prior.
If memory uses absolute child quality as edit credit, it can favor motifs applied to strong parents regardless of the edits' incremental value.
Residual memory addresses this ambiguity by subtracting an empirical child-quality reference for the parent-quality bucket.
A positive entry means that the motif's observed children outperform the historical reference for similarly scored parents, not that the motif is useful in every context.

This distinction is important for alpha mining because the base scorer already encodes factor quality, depth, and retrieval frequency through the selectable ledger, while admission separately enforces diversity.
The memory estimates empirical edit utility relative to the parent-quality reference. Reliable credit guides motif choice within the selected parent while the parent scorer remains unchanged.

\subsection{Bounded Deviation from the Search Prior}

The confidence-gated residual form provides a simple stability property.
For a finite search history, let $R_\Delta$ satisfy $|\Delta_t(g,m)|\leq R_\Delta$ for all observed cells, with $\lambda_t\leq\lambda_{\max}$, $c_t(g,m)\in[0,1]$, and $v_t(g,m)\in[0,1]$.
The deterministic memory contribution in Equation~\ref{eq:cgrm_score} satisfies
\begin{equation}
\begin{aligned}
    &\big|A_t(p,m)-\eta_{t,m}\\
    &\quad-\log\max(S_{\mathrm{ledger}}(p),\epsilon)\big|\\
    &\qquad\leq \lambda_{\max}R_\Delta+1.
\end{aligned}
\end{equation}
The bound applies to the deterministic memory contribution over the observed history, with Gaussian proposal noise represented separately by $\eta_{t,m}$.
During warmup motifs are sampled uniformly. With insufficient residual support the credit correction vanishes, although an independently supported failure penalty may remain active. Parent selection always uses the ledger.

\subsection{Asymmetric Process Veto}

No single failed edit is sufficient evidence for a reusable negative rule in noisy financial data.
The asymmetric process veto therefore activates only after a category-motif cell has at least five observations and an empirical failure rate of at least 0.8.
Flagged motifs receive a finite penalty and no residual promotion, whereas positive credit remains a soft preference.
This asymmetry targets repeated admission failures. Matched admission-failure controls evaluate its contribution under the same parent scorer and search budget.

\subsection{Interpretability of Self-Evolution}

Parametric self-evolution can improve generation behavior, but its learned search knowledge is difficult to inspect or edit.
Full-trajectory memory is inspectable, but retains information beyond the local transition statistics used here.
AlphaMemo exposes a middle ground.
Its memory table can be inspected as statements of the form:
\begin{tcolorbox}[colback=white!98!black,colframe=white!30!black,boxsep=1.1pt,top=6pt,fontupper=\itshape]
\centering
Under factor category $g$, realized edit motif $m$ has mean residual $\Delta$, confidence $c$, and empirical failure rate $\pi^-$.
\end{tcolorbox}
This makes the agent's self-evolution auditable by showing which edit behaviors are promoted, suppressed, or vetoed.

\section{Experiments}
\label{sec:experiments}

We examine final factor-pool utility and the contribution of search-level credit assignment.
The main comparison evaluates AlphaMemo's factor collections against predictive and formula-mining baselines.
Matched component controls separate residual credit, confidence attenuation, and admission-failure memory from the underlying parent scorer.

\begin{table}[b]
\centering
\small
\caption{Component roles in AlphaMemo.}
\label{tab:implemented_components}
\begin{tabular}{p{0.30\linewidth}p{0.56\linewidth}}
\toprule
\textbf{Component} & \textbf{Role} \\
\midrule
Search ledger & Selects parents using quality, depth, and retrieval frequency. \\
Structural motifs & Identify realized edits under the parent-factor context. \\
Residual SSPM & Estimates edit credit relative to historical child outcomes from similarly scored parents. \\
Confidence gates & Discount sparsely observed or noisy credit estimates. \\
Failure memory & Penalizes edits that repeatedly fail quality or redundancy admission. \\
\bottomrule
\end{tabular}
\end{table}

\begin{table*}[t!]
\centering
\caption{Performance comparison on CSI 500 and S\&P 500 under the 20-trading-day protocol. Higher is better for all columns, including signed MDD (closer to zero). AR and MDD are percentage points. Bold and underlined values denote the best and second-best results.}
\label{tab:main_comparison}
\setlength{\tabcolsep}{2.8pt}
\renewcommand{\arraystretch}{1.05}
\small
\resizebox{\textwidth}{!}{%
\begin{tabular}{lcccccccccccccc}
\toprule
\multirow{2}{*}{\textbf{Method}} & \multicolumn{7}{c}{\textbf{CSI 500}} & \multicolumn{7}{c}{\textbf{S\&P 500}} \\
\cmidrule(lr){2-8}\cmidrule(lr){9-15}
& \textbf{IC} & \textbf{ICIR} & \textbf{RankIC} & \textbf{RankICIR} & \textbf{AR(\%)} & \textbf{MDD(\%)} & \textbf{Sharpe}
& \textbf{IC} & \textbf{ICIR} & \textbf{RankIC} & \textbf{RankICIR} & \textbf{AR(\%)} & \textbf{MDD(\%)} & \textbf{Sharpe} \\
\midrule
Alpha158 & 0.0135 & 0.1224 & 0.0185 & 0.1580 & 6.66 & -38.72 & 0.3120 & 0.0135 & 0.1131 & \textbf{0.0099} & \textbf{0.0827} & \underline{16.21} & \textbf{-21.53} & 0.6847 \\
GP & 0.0154 & 0.1605 & \underline{0.0336} & \underline{0.3211} & 5.48 & \textbf{-26.24} & 0.2853 & -0.0055 & -0.0381 & -0.0110 & -0.0743 & 8.46 & -24.37 & 0.3492 \\
LightGBM & 0.0145 & 0.1693 & -0.0118 & -0.1136 & 4.58 & -34.37 & 0.2081 & 0.0140 & 0.1056 & 0.0053 & 0.0424 & 13.72 & -28.28 & 0.5273 \\
LSTM & \underline{0.0272} & \textbf{0.2831} & 0.0247 & 0.2714 & \textbf{9.47} & -32.79 & \underline{0.4119} & 0.0177 & 0.1054 & 0.0067 & 0.0403 & 15.94 & -27.88 & 0.5594 \\
AlphaGen & 0.0038 & 0.0319 & 0.0084 & 0.0725 & 2.09 & -44.63 & 0.0870 & 0.0146 & 0.1184 & 0.0039 & 0.0303 & 12.62 & -26.00 & 0.5669 \\
AlphaGPT & 0.0131 & 0.1404 & -0.0125 & -0.1135 & 3.30 & -36.78 & 0.1477 & 0.0144 & 0.1044 & -0.0017 & -0.0122 & 13.46 & -25.14 & 0.5693 \\
AlphaAgent & 0.0130 & 0.1252 & 0.0173 & 0.1636 & 7.07 & -37.16 & 0.3110 & \underline{0.0206} & \underline{0.1504} & 0.0049 & 0.0359 & 15.56 & \underline{-24.20} & 0.6500 \\
AlphaSAGE & 0.0038 & 0.0361 & 0.0087 & 0.1011 & -0.28 & -45.18 & -0.0039 & 0.0132 & 0.1298 & 0.0057 & \underline{0.0667} & 14.90 & -25.14 & \underline{0.6876} \\
\midrule
\rowcolor{oursrow}
AlphaMemo & \textbf{0.0290} & \underline{0.2818} & \textbf{0.0382} & \textbf{0.4033} & \underline{8.83} & \underline{-31.21} & \textbf{0.4169} & \textbf{0.0277} & \textbf{0.2097} & \underline{0.0072} & 0.0552 & \textbf{17.93} & -24.45 & \textbf{0.7810} \\
\bottomrule
\end{tabular}}
\end{table*}

\subsection{Experimental Setup}

\paragraph{Datasets and splits.}
We use CSI 500 and S\&P 500 from Qlib~\citep{yang2020qlib}, covering Chinese A-shares and U.S. large caps.
Training and online search use 2016--2020, downstream validation uses 2021, and evaluation uses 2022--2025, ending on 2025-12-26.
The target is a 20-trading-day forward return. Signal dates whose labels cross a split boundary are removed.
SSPM is updated only with search-period feedback and remains frozen during downstream validation and evaluation.
A shared reference scheduler is selected using validation ICIR averaged equally over both markets and two search seeds.
Three additional search seeds confirm this scheduler for the matched component controls.
Table~\ref{tab:implemented_components} summarizes the corresponding component roles.

\paragraph{Metrics.}
We report IC, ICIR, RankIC, and RankICIR for cross-sectional prediction, and AR, MDD, and Sharpe for the realized TopK portfolio.
Let $s_{i,t}$ denote a factor or model score for asset $i$ at date $t$, and let $r^{(20)}_{i,t}$ be its next-close-aligned 20-trading-day return.
The daily IC measures cross-sectional predictive association:
\begin{equation}
\label{eq:main_ic}
    \mathrm{IC}_t=\mathrm{corr}_{i}\big(s_{i,t},r^{(20)}_{i,t}\big).
\end{equation}
The reported IC is its temporal mean. ICIR normalizes this mean by its temporal standard deviation:
\begin{equation}
\label{eq:main_icir}
    \mathrm{ICIR}=\frac{\mathbb{E}_t[\mathrm{IC}_t]}
    {\mathrm{Std}_t[\mathrm{IC}_t]+\epsilon}.
\end{equation}
Here $\epsilon$ is a numerical stabilizer. Rank-based metrics measure monotonic predictive association and its temporal stability:
\begin{align}
\label{eq:main_rankic}
    \mathrm{RankIC}_t={}&\mathrm{corr}_{i}\big(\mathrm{rank}\,s_{i,t},
    \mathrm{rank}\,r^{(20)}_{i,t}\big),\\
\label{eq:main_rankicir}
    \mathrm{RankICIR}={}&\frac{\mathbb{E}_t[\mathrm{RankIC}_t]}
    {\mathrm{Std}_t[\mathrm{RankIC}_t]+\epsilon}.
\end{align}
Ranks are computed across assets at each date, and the reported RankIC is the temporal mean.
Portfolio metrics use gross daily returns rather than benchmark-excess information ratios.
Appendix~\ref{app:metrics} provides the full metric definitions and return alignment, and Appendix~\ref{app:matched_credit} gives per-control dispersion.

\paragraph{Baselines and Implementation.}
The baseline set includes the expert factor library Alpha158~\citep{yang2020qlib}, GP~\citep{chen2021gp}, LightGBM~\citep{ke2017lightgbm}, and LSTM~\citep{schmidhuber1997long}.
Formula-mining systems comprise AlphaGen~\citep{yu2023alphagen}, AlphaGPT~\citep{wang2025alpha}, AlphaAgent~\citep{tang2025alphaagent}, and AlphaSAGE~\citep{chen2025alphasage}.
Released symbolic operators are retained when exporting external pools to the common downstream evaluator.
Internal controls are reported separately from the external baselines.

AlphaMemo uses DeepSeek v4 Flash for 500 search attempts in batches of five and exports at most 50 factors.
All internal controls share the generator, parent scorer, budget, admission rule, and downstream evaluator.
Appendix~\ref{app:more_exp} details the baseline interfaces, repetition counts, and evaluation settings, followed by the scheduler, admission thresholds, completion handling, and factor selection.

\begin{table*}[t]
\centering
\small
\caption{Matched ablation study on CSI 500 with the shared scheduler. Values are mean $\pm$ sample standard deviation over three search seeds. Higher is better for all columns, including signed MDD (closer to zero). AR and MDD are percentage points. The best means are bolded.}
\label{tab:credit-controls-csi500}
\label{tab:qlib_ablation}
\setlength{\tabcolsep}{3pt}
\resizebox{\textwidth}{!}{%
\begin{tabular}{lccccccc}
\toprule
\textbf{Method} & \textbf{IC} & \textbf{ICIR} & \textbf{RankIC} & \textbf{RankICIR} & \textbf{AR(\%)} & \textbf{MDD(\%)} & \textbf{Sharpe} \\
\midrule
\textbf{AlphaMemo} & $\mathbf{0.0274} \pm 0.0092$ & $\mathbf{0.2698} \pm 0.0559$ & $\mathbf{0.0383} \pm 0.0203$ & $\mathbf{0.3604} \pm 0.1206$ & $\mathbf{8.01} \pm 3.47$ & $\mathbf{-28.24} \pm 5.85$ & $\mathbf{0.3994} \pm 0.1816$ \\
\quad w/o process memory & $0.0192 \pm 0.0044$ & $0.1937 \pm 0.0297$ & $0.0333 \pm 0.0091$ & $0.3126 \pm 0.0640$ & $4.65 \pm 1.42$ & $-30.01 \pm 3.01$ & $0.2380 \pm 0.0770$ \\
\quad Absolute credit & $0.0223 \pm 0.0106$ & $0.2468 \pm 0.0821$ & $0.0284 \pm 0.0134$ & $0.3142 \pm 0.1092$ & $6.27 \pm 2.90$ & $-32.47 \pm 4.70$ & $0.2956 \pm 0.1349$ \\
\quad w/o confidence gate & $0.0192 \pm 0.0027$ & $0.1857 \pm 0.0137$ & $0.0299 \pm 0.0105$ & $0.2911 \pm 0.0739$ & $7.66 \pm 1.75$ & $-29.98 \pm 1.62$ & $0.3788 \pm 0.0959$ \\
\quad w/o failure penalty & $0.0202 \pm 0.0088$ & $0.2047 \pm 0.0563$ & $0.0290 \pm 0.0143$ & $0.2900 \pm 0.1234$ & $4.94 \pm 1.71$ & $-31.35 \pm 4.74$ & $0.2429 \pm 0.1000$ \\
\bottomrule
\end{tabular}}
\end{table*}

\begin{figure*}[t!]
\centering
\includegraphics[width=0.95\textwidth]{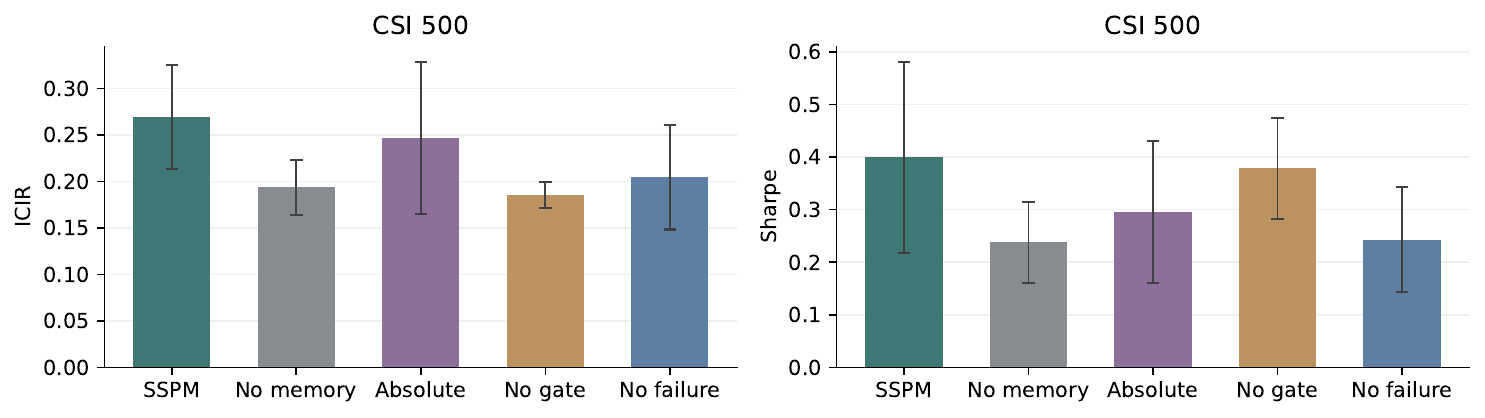}
\caption{Matched credit-assignment controls on CSI 500. Bars show three-seed means of ICIR and Sharpe. Error bars show sample standard deviations.}
\label{fig:memory_calibration}
\end{figure*}

\begin{table}[t]
\centering
\small
\setlength{\tabcolsep}{3pt}
\caption{Memory-scheduler sensitivity. $W$ denotes warmup and $\lambda$ residual weight. Entries are means over two search seeds. Full dispersion is reported in Appendix~\ref{app:scheduler_sensitivity}.}
\label{tab:memory_weight_sensitivity}
\begin{tabular}{cccccc}
\toprule
\multirow{2}{*}{$W$} & \multirow{2}{*}{$\lambda$} & \multicolumn{2}{c}{\textbf{CSI 500}} & \multicolumn{2}{c}{\textbf{S\&P 500}} \\
\cmidrule(lr){3-4}\cmidrule(lr){5-6}
& & ICIR & Sharpe & ICIR & Sharpe \\
\midrule
120 & 0.04 & 0.2624 & 0.3936 & 0.2288 & 0.7533 \\
180 & 0.04 & 0.2509 & 0.2969 & 0.1866 & 0.8049 \\
240 & 0.04 & 0.0135 & 0.1882 & 0.2104 & 0.7533 \\
\midrule
180 & 0.02 & 0.1514 & 0.2854 & 0.1083 & 0.6163 \\
180 & 0.08 & 0.3197 & 0.2789 & 0.1252 & 0.6803 \\
\bottomrule
\end{tabular}
\end{table}

\subsection{Performance Comparison}

Table~\ref{tab:main_comparison} compares cross-sectional prediction and portfolio utility.
On CSI 500, AlphaMemo exceeds all four traditional baselines in IC, RankIC, RankICIR and Sharpe.
The gains in both IC and RankIC indicate better cross-sectional prediction under linear and rank-based measures.
On S\&P 500, AlphaMemo surpasses Alpha158 and LSTM in IC, ICIR, AR and Sharpe.
The accompanying AR and Sharpe gains show that the predictive advantage also translates into portfolio returns under the common construction rule.

\subsection{Ablation Study}

The ablations use a fixed shared scheduler and matched search seeds, rather than the market-specific main settings.
Table~\ref{tab:qlib_ablation} and Figure~\ref{fig:memory_calibration} show that full SSPM has the strongest mean predictive and portfolio results among the CSI 500 controls, with higher ICIR than no memory in every seed.
Its advantage over absolute child-quality memory supports adjusting edit credit for parent quality, instead of reinforcing edits mainly because they originate from strong parents.
Removing confidence attenuation reduces mean predictive stability, consistent with discounting poorly supported residual estimates before reusing them.
Removing failure memory weakens both predictive and portfolio performance, suggesting that admission feedback complements numerical credit by discouraging recurrent low-quality or redundant proposals.

S\&P 500 shows mixed effects, with higher mean Sharpe but lower ICIR than no memory. Appendix~\ref{app:matched_credit} reports these controls and paired results. The shared parameters are listed in Table~\ref{tab:appendix_operating_points} of Appendix~\ref{app:more_exp}.

\subsection{Memory-Scheduler Sensitivity}

Table~\ref{tab:memory_weight_sensitivity} varies warmup at $\lambda=0.04$ and residual weight at $W=180$, fixing failure weight 1.0, admission threshold 0.10, and batch size five.
Longer warmup reduces CSI 500 performance, while S\&P 500 favors different timings for ICIR and Sharpe. Stronger residual weighting improves CSI 500 ICIR but not Sharpe, and S\&P 500 favors the intermediate weight.
Delaying reuse leaves fewer guided attempts, while stronger corrections need not improve portfolio outcomes.
Appendix~\ref{app:scheduler_sensitivity} reports the full sweep and variability.

\section{Conclusion}
\label{sec:conclusion}
We presented AlphaMemo, a self-evolving alpha-mining agent that organizes Structured Search-Process Memory around search-level credit assignment. Realized formula edits define the experience, parent-adjusted outcomes provide relative credit, and confidence and admission-failure evidence control its reuse. This design targets inherited predictive signals, noisy financial feedback, and redundancy constraints in executable alpha search. Exploratory market-specific comparisons show predictive and portfolio advantages over multiple factor-mining baselines, and matched three-seed controls support SSPM's contribution on CSI 500. These findings highlight compact, reference-adjusted edit memory as an inspectable mechanism for reusing alpha-search experience.



\section*{Limitations}
AlphaMemo studies offline discovery with executable formulas and standardized search feedback.
Memory estimates depend on repeated observations and may vary with market conditions. Extending the method to broader formula spaces and live trading requires further validation and operational safeguards.

\section*{Ethical Considerations}
\label{sec:ethics}
All experiments use publicly available datasets and benchmarks. No human subjects or sensitive data are involved. No direct negative societal impacts are identified.

\section*{Acknowledgments}

HY was supported in part by a JD Technology Research Scholarship in Artificial Intelligence.

\bibliography{reference}

\appendix

\section{Algorithmic Details and Derivation}
\label{app:algorithm_theory}

\subsection{AlphaMemo Search Procedure}

Algorithm~\ref{alg:alphamemo_appendix} summarizes the full search loop.
The base scorer retrieves promising parent factors from the selectable ledger, while Structured Search-Process Memory (SSPM) supplies a confidence-gated residual correction over structural edit motifs observed in previous parent-child transitions.
Parents are selected before motif scoring, so SSPM controls editing choices without reranking the parent list.

\subsection{Residual Process Memory}

Let $p$ be a parent formula and $c$ the child produced after requesting motif $m$.
For the selectable ledger, define the standardized quality
\begin{equation}
Z_Q(p)=\frac{Q(p)-\overline Q}{s_Q+\epsilon}.
\end{equation}
The parent score is
\begin{equation}
\begin{split}
S_{\mathrm{ledger}}(p)={}&
\operatorname{sigmoid}\!\left(Z_Q(p)\right)
\cdot(1-d_{\mathrm{depth}})^{D(p)}\\
&\cdot(1-d_{\mathrm{use}})^{\max(U(p)-U_0,0)},
\end{split}
\end{equation}
where $D(p)$ and $U(p)$ are depth and retrieval count.

SSPM does not store the absolute quality of the child as its edit effect.
Parent quality is assigned to a low, medium, or high bucket $b_q(p)$, and the baseline is the mean quality of prior children whose parents occupy that bucket:
let $\mathcal{H}_b=\mathcal{H}_t(b_q(p))$ contain previously evaluated, distinct valid children, and for a nonempty bucket define
\begin{equation}
\overline Q_b=|\mathcal{H}_b|^{-1}
\sum_{c_i\in\mathcal{H}_b}Q(c_i).
\end{equation}
Then
\begin{equation}
\widehat{Q}_{\mathrm{ledger}}(p)=
\begin{cases}
\overline Q_b, & |\mathcal{H}_b|>0,\\
Q(p), & \text{otherwise}.
\end{cases}
\end{equation}
The stored residual is $\delta(p,c)=Q(c)-\widehat{Q}_{\mathrm{ledger}}(p)$, computed before adding the current child to the reference history.
The parent category supplies both the query and update context. Invalid and syntactically duplicate children remain in the audit record but do not enter credit or failure statistics.

For a memory key $(g,m)$, SSPM maintains empirical mean and population variance:
\begin{align}
    \mu_{g,m}&=\frac{1}{n_{g,m}}\sum_i \delta_i,\\
    \sigma_{g,m}^2&=\frac{1}{n_{g,m}}\sum_i(\delta_i-\mu_{g,m})^2.
\end{align}
Let $\hat p_{g,m}=(n^+_{g,m}+1)/(n_{g,m}+2)$ and let $H_2$ denote binary entropy. The confidence used by the reported implementation is
\begin{equation}
\begin{split}
c_t(g,m)={}&\min\left(1,\frac{n_{g,m}}{n_{\mathrm{conf}}}\right)
\left(1-\frac{H_2(\hat p_{g,m})}{2}\right)\\
&\cdot\max\left(0,1-\frac{\sigma_{g,m}}
{|\mu_{g,m}|+0.03+\epsilon}\right),
\end{split}
\end{equation}
with $c_t(g,m)=0$ for fewer than two observations.
This form jointly discounts low-support, inconsistent, and high-variance cells.

\begin{algorithm}[t]
\caption{AlphaMemo factor evolution loop.}
\label{alg:alphamemo_appendix}
\small
\begin{algorithmic}[1]
\Require seed parents, budget $B$, batch size $N$, warmup $T$, random exploration probability $\rho$, thresholds $\tau_q,\tau_d,n_v,\tau_v$
\Ensure admitted factor pool $\mathcal{P}_B$
\State Initialize $\mathcal{G}$ with valid seed parents, $\mathcal{P}\leftarrow\varnothing$, and $\mathcal{M}\leftarrow\mathcal{M}_0$
\For{$t=0,N,2N,\ldots$ while $t<B$}
    \State Retrieve up to $\min(N,B-t)$ parents using $S_{\mathrm{ledger}}$
    \For{each selected parent $p$}
        \If{$t<T$ or a random exploration draw is below $\rho$}
            \State Sample a requested motif $m$ uniformly
        \Else
            \State Sample four motifs and query SSPM for $(g(p),m)$
            \State Select $m$ by Equation~\ref{eq:cgrm_score}, including failure penalties and proposal noise
        \EndIf
        \State Generate a child from $p$, its category, requested $m$, and executable-operator constraints
    \EndFor
    \For{each generated parent--child pair $(p,c)$ in the batch}
        \State Evaluate $c$ on the search period and record the attempt
        \If{$c$ is invalid or syntactically duplicate}
            \State \textbf{continue}
        \EndIf
        \State Determine admission status $s$ using quality and correlation constraints
        \State Derive realized motif $m(p,c)$ and compute $\delta(p,c)$ using the previous reference history
        \State Update credit and failure statistics at $(g(p),m(p,c))$, then update the reference history
        \If{$s$ is admitted}
            \State Add $c$ and $(p,c)$ to the ledger and admit $c$ into $\mathcal{P}$
        \EndIf
    \EndFor
\EndFor
\State \Return $\mathcal{P}$
\end{algorithmic}
\end{algorithm}

\subsection{Asymmetric Process Veto}

Positive residual evidence enters only through the soft correction in Equation~\ref{eq:cgrm_score}.
Failure handling uses the empirical rate $\pi^-_{g,m}=n^-_{g,m}/n_{g,m}$ and flags an action when
\begin{equation}
    \mathrm{Veto}(g,m)=
    \mathbb{I}\left[n_{g,m}\ge n_v
    \ \wedge\
    \pi^-_{g,m}\ge\tau_v\right].
\end{equation}
The reported runs use $n_v=5$ and $\tau_v=0.8$.
For fewer than five observations the penalty is zero. Thereafter a flagged motif has severity $\pi^-_{g,m}$ and no residual correction. Other motifs have severity $\pi^-_{g,m}/2$. The penalty $v_t(h)$ scales this severity by the configured failure-memory weight.
The penalty is finite, so flagged motifs remain available when all alternatives score worse or through random exploration.
These counts describe failed factor admission among distinct valid edits, rather than generator syntax errors or a claim that negative patterns persist across regimes.

\section{Evaluation Metrics}
\label{app:metrics}

We follow the standard factor-mining convention of reporting both predictive correlations and portfolio-level utility.
For a factor $f$, let $f_{i,t}$ denote the score of asset $i$ at date $t$, and let $r^{(h)}_{i,t}$ be its next-close-aligned future holding-period return.
The main experiments use $h=20$ trading days and define the close-to-close forward return as
\begin{equation}
    r^{(h)}_{i,t}=\frac{C_{i,t+1+h}}{C_{i,t+1}}-1,
\end{equation}
where $C_{i,t}$ is the adjusted close price.
In the Qlib implementation, this corresponds to the next-tradable-close aligned label
\texttt{Ref(\$close,-21) / Ref(\$close,-1) - 1}.
The daily Information Coefficient is the cross-sectional Pearson correlation
\begin{equation*}
    \mathrm{IC}_t(f)=
    \mathrm{corr}_{i}\big(f_{i,t}, r^{(h)}_{i,t}\big).
\end{equation*}
RankIC replaces raw scores and returns with their cross-sectional ranks:
\begin{equation*}
    \mathrm{RankIC}_t(f)=
    \mathrm{corr}_{i}\big(\mathrm{rank}(f_{i,t}), \mathrm{rank}(r^{(h)}_{i,t})\big).
\end{equation*}
The reported IC and RankIC are temporal means over the test period.
Their information ratios measure stability:
\begin{align*}
    \mathrm{ICIR}(f) &=
    \frac{\mathbb{E}_t[\mathrm{IC}_t(f)]}
    {\mathrm{Std}_t[\mathrm{IC}_t(f)]+\epsilon},\\
    \mathrm{RankICIR}(f) &=
    \frac{\mathbb{E}_t[\mathrm{RankIC}_t(f)]}
    {\mathrm{Std}_t[\mathrm{RankIC}_t(f)]+\epsilon}.
\end{align*}

For portfolio metrics, AR is annualized portfolio return, MDD is maximum drawdown, and Sharpe is the annualized return-to-volatility ratio of the backtested strategy.
For realized gross daily returns $u_t$, the implementation uses $\mathrm{AR}=252\,\overline{u}$ and $\mathrm{Sharpe}=\sqrt{252}\,\overline{u}/\mathrm{Std}(u)$.
MDD is the minimum ratio of compounded NAV to its running peak, minus one, including initial NAV one in the peak calculation.
In the main table, AR and MDD are reported in percentage points.

\paragraph{Search efficiency.}
For fixed-budget diagnostics, we count effective rather than raw generated factors.
Given the current admitted pool $\mathcal{P}$, define the maximum absolute correlation between a candidate $f$ and the existing pool as
\begin{equation*}
    \rho_{\max}(f,\mathcal{P}) =
    \max_{g\in\mathcal{P}} |\rho(f,g)|.
\end{equation*}
An effective factor is an executable expression that passes the search-period quality threshold, satisfies the expression-validity checks, and is not redundant with the current pool:
\begin{align*}
    \mathrm{Eff}(f,\mathcal{P})
    =
    \mathbb{I}\{
    \mathrm{valid}(f)
    \wedge | \mathrm{ICIR}_{\mathrm{search}}(f)| \geq \tau_q \notag\\
    \wedge \rho_{\max}(f,\mathcal{P}) \leq \tau_d
    \}.
\end{align*}
The fixed-budget discovery yield is
\begin{equation*}
    N_{\mathrm{eff}}(B)=\sum_{b=1}^{B}\mathrm{Eff}(f_b,\mathcal{P}_{b-1}),
\end{equation*}
where $B$ is the generation budget.

\paragraph{Alpha decay.}
To measure whether a factor pool remains useful under non-stationarity, we compute annual IC and RankIC on held-out years.
For a factor pool $\mathcal{P}$, we first form a normalized pool signal
\begin{equation*}
    s_{i,t}=\frac{1}{|\mathcal{P}|}\sum_{f\in\mathcal{P}}
    \frac{f_{i,t}-\mu_t(f)}{\sigma_t(f)+\epsilon},
\end{equation*}
where $\mu_t(f)$ and $\sigma_t(f)$ are cross-sectional statistics at date $t$.
Annual alpha decay is then reported as
\begin{equation*}
    \mathrm{IC}^{(y)}(\mathcal{P})=
    \mathbb{E}_{t:\mathrm{year}(t)=y}
    \left[\mathrm{corr}_{i}(s_{i,t}, r^{(h)}_{i,t})\right],
\end{equation*}
with an analogous definition for annual RankIC.
This diagnostic asks whether the evolved factor pool retains predictive content across calendar years rather than only on the aggregate test window.

\section{Additional Experiments}
\label{app:more_exp}

\subsection{Matched Search-Level Credit Assignment}
\label{app:matched_credit}

\paragraph{Question and memory design.}
A useful parent factor can yield a strong child even when the requested edit contributes little.
SSPM therefore asks which realized edits deserve credit beyond the quality expected from their parents.
This is search-level credit assignment: executable parent--child changes provide the unit of experience, search-period predictive feedback provides its value, and accumulated evidence guides subsequent edit proposals.
The credit target is child $|\mathrm{ICIR}|$ minus the mean quality of previously evaluated, distinct valid children in the same parent-quality bucket, with the parent quality as the empty-bucket reference.
It is an empirical reference adjustment, not an identified causal effect of the edit.
Each child receives direct predictive feedback. Credit is assigned to its local edit, rather than by propagating a terminal reward through an entire search trajectory.
Both retrieval and updates use the parent category, and the realized motif is computed from parsed expression profiles.
Invalid and duplicate attempts are recorded separately and do not update numerical credit or empirical failure counts.

\paragraph{Shared configuration and controls.}
One configuration is selected using signed downstream validation ICIR, averaged equally over the two markets and selection seeds 0 and 1.
The selected configuration uses a 180-attempt warmup, memory weight 0.04, failure-memory weight 1.0, four sampled motifs, and random motif probability 0.35.
The main comparison instead uses market-specific settings and seeds 0, 1, and 2. The ablations below use the shared setting and separate confirmation seeds.
Confirmation uses search seeds 2, 3, and 4, distinct from selection seeds, with 500 attempts per run, batches of five, an admission threshold of $|\mathrm{ICIR}|\geq0.10$, and a correlation threshold of 0.70.
The no-memory control disables both credit corrections and empirical failure penalties.
The absolute-credit control substitutes child $|\mathrm{ICIR}|$ for residual credit, while the no-confidence control uses the empirical residual mean without confidence attenuation.
The no-veto control retains residual credit and confidence attenuation but removes admission-failure penalties and veto flags from scheduling.
All controls retain the same parent scorer, motif sampling, proposal noise, generator settings, admission rule, and evaluation procedure.

\paragraph{Evaluation and generation.}
Search feedback comes from 2016--2020, downstream validation from 2021, and reported evaluation from 2022--2025.
The last 21 signal dates of each split are removed so that the 20-trading-day forward label remains within that split.
Downstream selection ranks valid candidates using search-period $|\mathrm{ICIR}|$, requires at least 80 valid metric days, and retains up to 50 factors.
Generation uses \texttt{deepseek/deepseek-v4-flash}, an output limit of 1,024 tokens, disabled reasoning, and recorded completion status.
Malformed completions receive bounded retries. Unrecoverable candidate-level formatting failures count against the search budget without a substitute formula or a memory update, and sustained failures stop the run.
Portfolio metrics use realized gross returns, with MDD computed from compounded net asset value.

\begin{table*}[t]
\centering
\small
\caption{Matched ablation study on S\&P 500 with the shared scheduler. Values are mean $\pm$ sample standard deviation over three search seeds. Higher is better for all columns, including signed MDD (closer to zero). AR and MDD are percentage points.}
\label{tab:credit-controls-sp500}
\setlength{\tabcolsep}{3pt}
\resizebox{\textwidth}{!}{%
\begin{tabular}{lccccccc}
\toprule
\textbf{Method} & \textbf{IC} & \textbf{ICIR} & \textbf{RankIC} & \textbf{RankICIR} & \textbf{AR(\%)} & \textbf{MDD(\%)} & \textbf{Sharpe} \\
\midrule
\textbf{AlphaMemo} & $0.0204 \pm 0.0126$ & $0.1599 \pm 0.1029$ & $0.0016 \pm 0.0046$ & $0.0145 \pm 0.0370$ & $16.51 \pm 4.03$ & $-24.93 \pm 1.16$ & $0.7071 \pm 0.2081$ \\
\quad w/o process memory & $0.0260 \pm 0.0014$ & $0.1903 \pm 0.0163$ & $0.0050 \pm 0.0029$ & $0.0348 \pm 0.0204$ & $16.26 \pm 0.88$ & $-26.50 \pm 0.71$ & $0.6847 \pm 0.0255$ \\
\quad Absolute credit & $0.0213 \pm 0.0052$ & $0.1571 \pm 0.0376$ & $0.0024 \pm 0.0049$ & $0.0164 \pm 0.0343$ & $15.85 \pm 1.79$ & $-25.88 \pm 2.85$ & $0.6735 \pm 0.0666$ \\
\quad w/o confidence gate & $0.0199 \pm 0.0208$ & $0.1520 \pm 0.1610$ & $0.0009 \pm 0.0095$ & $0.0057 \pm 0.0745$ & $15.95 \pm 2.48$ & $-24.69 \pm 0.75$ & $0.6902 \pm 0.1115$ \\
\quad w/o failure penalty & $0.0258 \pm 0.0064$ & $0.1928 \pm 0.0472$ & $0.0028 \pm 0.0044$ & $0.0200 \pm 0.0322$ & $17.06 \pm 2.25$ & $-24.92 \pm 0.49$ & $0.7440 \pm 0.0738$ \\
\bottomrule
\end{tabular}}
\end{table*}


\begin{figure*}[t]
\centering
\includegraphics[width=0.95\textwidth]{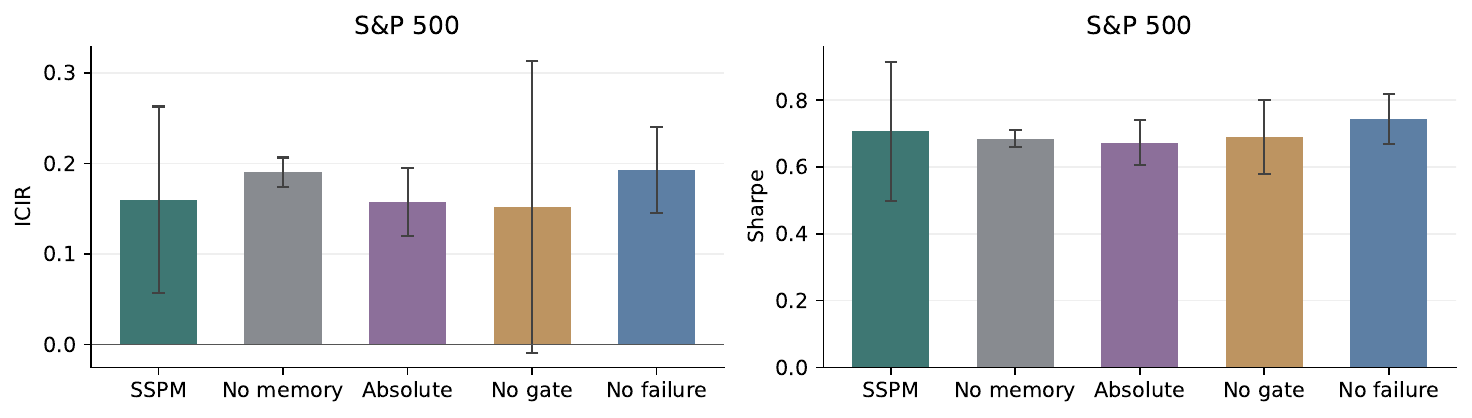}
\caption{Matched credit-assignment controls on S\&P 500. Bars show three-seed means of ICIR and Sharpe. Error bars show sample standard deviations.}
\label{fig:matched_credit_sp500}
\end{figure*}

\begin{figure*}[t]
\centering
\includegraphics[width=0.95\textwidth]{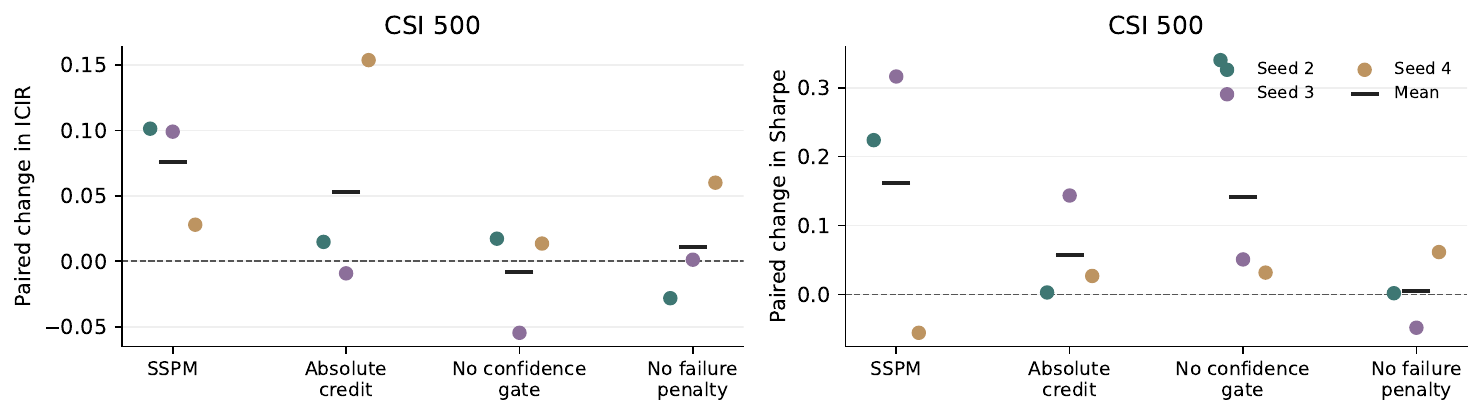}
\caption{Paired effects of credit-assignment controls on CSI 500 relative to the no-memory control. Each point compares runs with the same search seed. Black marks denote three-seed means.}
\label{fig:paired_credit_csi500}
\end{figure*}

\paragraph{Paired evidence.}
Table~\ref{tab:credit-controls-csi500} and Figure~\ref{fig:memory_calibration} in the main text report the matched CSI 500 controls. Table~\ref{tab:credit-controls-sp500} and Figure~\ref{fig:matched_credit_sp500} report the corresponding S\&P 500 results.
Figure~\ref{fig:paired_credit_csi500} displays same-seed CSI 500 differences relative to the no-memory control.

Full SSPM improves ICIR over the no-memory control in all three seeds, with paired differences of 0.1014, 0.0991, and 0.0280.
Its mean ICIR is 0.2698, compared with 0.1937 without memory, 0.2468 with absolute credit, and 0.1857 without confidence attenuation.
The absolute-credit comparison favors residual credit in two of three seeds, so the average improvement should be read together with the reported dispersion.

Table~\ref{tab:credit-controls-sp500} reports the corresponding S\&P 500 controls.
Full SSPM has mean ICIR 0.1599 versus 0.1903 without memory, with paired differences 0.0290, 0.0234, and $-0.1437$.
Its mean Sharpe is 0.7071 versus 0.6847 without memory, but the third seed also reverses the Sharpe comparison.
Thus the current shared configuration provides stronger evidence on CSI 500 than on S\&P 500. The complete matched results do not establish uniform cross-market predictive gains.

Removing failure memory yields mean ICIR 0.2047 on CSI 500 and 0.1928 on S\&P 500, with mean Sharpe 0.2429 and 0.7440, respectively.
Thus failure suppression is favorable in the CSI 500 mean comparison but not in the S\&P 500 predictive or Sharpe comparison.
These controls examine credit estimation, confidence attenuation, and failure suppression. They do not independently establish the contribution of the structural motif extractor.

\subsection{Memory-Scheduler Sensitivity}
\label{app:historical_calibration}

\begin{figure*}[t!]
\centering
\includegraphics[width=0.9\textwidth]{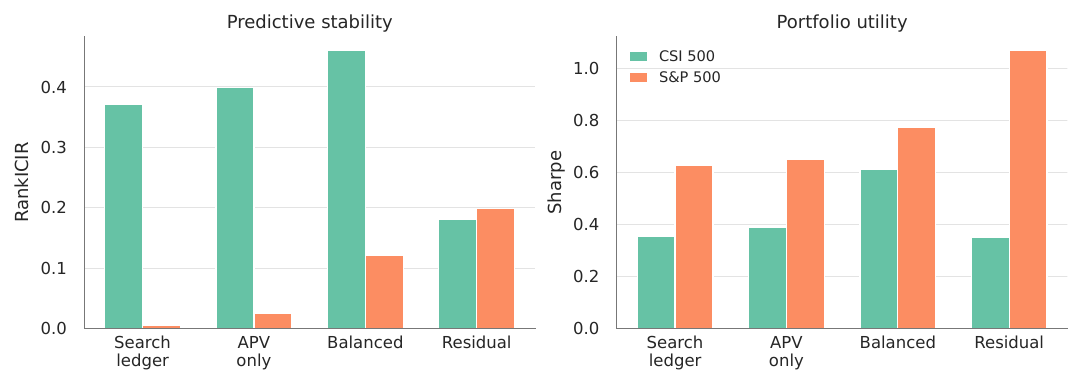}
\caption{Memory calibration under alternative scheduling settings on CSI 500 and S\&P 500. Bars report single-run values.}
\label{fig:historical_memory_calibration}
\end{figure*}

Figure~\ref{fig:historical_memory_calibration} and Table~\ref{tab:historical_memory_sensitivity} examine AlphaMemo under alternative memory schedulers.
The search-ledger control removes process memory and APV-only retains failure suppression. Balanced scheduling and residual scheduling denote the two memory schedulers in these diagnostics, not the market-specific settings in Table~\ref{tab:main_comparison}.
These single-run diagnostics retain signal dates at split boundaries, whereas the matched component study removes dates whose forward labels cross those boundaries.
They assess scheduling sensitivity under that setting. Comparisons are made within each study rather than pooling results across evaluation settings.
Appendix~\ref{app:matched_credit} reports the three-seed component controls and their dispersion.

\begin{table*}[t!]
\centering
\small
\caption{Sensitivity to memory scheduling on CSI 500 and S\&P 500. AR is reported in percentage points.}
\label{tab:historical_memory_sensitivity}
\setlength{\tabcolsep}{4pt}
\resizebox{0.8\textwidth}{!}{%
\begin{tabular}{lcccccccc}
\toprule
\multirow{2}{*}{\textbf{Variant}} & \multicolumn{4}{c}{\textbf{CSI 500}} & \multicolumn{4}{c}{\textbf{S\&P 500}} \\
\cmidrule(lr){2-5}\cmidrule(lr){6-9}
& \textbf{ICIR} & \textbf{RankICIR} & \textbf{AR(\%)} & \textbf{Sharpe} & \textbf{ICIR} & \textbf{RankICIR} & \textbf{AR(\%)} & \textbf{Sharpe} \\
\midrule
Search-ledger only & 0.2313 & 0.3707 & 7.00 & 0.3536 & 0.1414 & 0.0044 & 13.80 & 0.6282 \\
Balanced scheduling & 0.3462 & 0.4597 & 11.63 & 0.6109 & 0.2406 & 0.1207 & 17.07 & 0.7743 \\
Weaker memory & 0.1455 & 0.2886 & 5.80 & 0.3011 & 0.1001 & -0.0299 & 17.09 & 0.7393 \\
Stronger memory & 0.2208 & 0.3454 & 11.57 & 0.5646 & 0.1579 & 0.0402 & 16.20 & 0.7105 \\
APV-only memory & 0.2405 & 0.3993 & 7.72 & 0.3875 & 0.1718 & 0.0243 & 15.99 & 0.6493 \\
Residual scheduling & 0.1104 & 0.1808 & 6.97 & 0.3511 & 0.3434 & 0.1984 & 23.65 & 1.0672 \\
\bottomrule
\end{tabular}}
\end{table*}

\subsection{Scheduler and Admission Sensitivity}
\label{app:scheduler_sensitivity}

Table~\ref{tab:scheduler_sensitivity} reports all 15 evaluated configurations on each market.
Each setting uses 500 search attempts and the same downstream evaluation protocol.
The table holds search seeds fixed at 0 and 1 across settings, giving two complete runs per entry.
Additional repeats of the main configurations are summarized separately in the main comparison.
These two-run estimates characterize sensitivity, with sample standard deviations for ICIR and Sharpe indicating search variability.

\begin{table*}[t]
\centering
\footnotesize
\setlength{\tabcolsep}{2.2pt}
\renewcommand{\arraystretch}{1.08}
\caption{Sensitivity to scheduler and admission settings on CSI 500 and S\&P 500. Values are means over two search seeds. ICIR and Sharpe also show sample standard deviations. $W$ is warmup, $\lambda$ is residual weight, $\gamma$ is failure weight, $\tau_q$ is the admission quality threshold, and $N$ is batch size. AR and MDD are percentage points.}
\label{tab:scheduler_sensitivity}
\begin{tabular}{cccccccccccc}
\toprule
\multicolumn{12}{c}{\textbf{CSI 500}} \\
\midrule
$W$ & $\lambda$ & $\gamma$ & $\tau_q$ & $N$ & IC & ICIR & RankIC & RankICIR & AR(\%) & MDD(\%) & Sharpe \\
\midrule
120 & 0.04 & 1 & 0.1 & 5 & 0.0197 & $0.2624 \pm 0.1958$ & 0.0312 & 0.4167 & 8.32 & -28.69 & $0.3936 \pm 0.0259$ \\
120 & 1 & 0.05 & 0.1 & 5 & 0.0214 & $0.2415 \pm 0.0406$ & 0.0298 & 0.3214 & 7.12 & -27.42 & $0.3617 \pm 0.1476$ \\
180 & 0.02 & 1 & 0.1 & 5 & 0.0164 & $0.1514 \pm 0.1305$ & 0.0299 & 0.3026 & 5.88 & -27.00 & $0.2854 \pm 0.0172$ \\
180 & 0.04 & 0.05 & 0.1 & 5 & 0.0191 & $0.2246 \pm 0.1336$ & 0.0345 & 0.3778 & 5.56 & -30.11 & $0.2799 \pm 0.1827$ \\
180 & 0.04 & 1 & 0.02 & 5 & 0.0285 & $0.2594 \pm 0.1575$ & 0.0450 & 0.3931 & 7.84 & -34.39 & $0.3726 \pm 0.2387$ \\
180 & 0.04 & 1 & 0.02 & 10 & 0.0192 & $0.1891 \pm 0.0712$ & 0.0293 & 0.2804 & 5.05 & -30.09 & $0.2535 \pm 0.1007$ \\
180 & 0.04 & 1 & 0.1 & 5 & 0.0250 & $0.2509 \pm 0.1257$ & 0.0388 & 0.4239 & 6.38 & -32.78 & $0.2969 \pm 0.1968$ \\
180 & 0.08 & 1 & 0.1 & 5 & 0.0313 & $0.3197 \pm 0.0208$ & 0.0302 & 0.3080 & 6.03 & -31.82 & $0.2789 \pm 0.0093$ \\
180 & 0.4 & 0.05 & 0.1 & 5 & 0.0062 & $0.0848 \pm 0.2387$ & 0.0050 & 0.0686 & 7.15 & -34.08 & $0.3261 \pm 0.1439$ \\
180 & 1 & 0.05 & 0.1 & 5 & 0.0148 & $0.1866 \pm 0.0148$ & 0.0197 & 0.2228 & 5.27 & -29.96 & $0.2602 \pm 0.1167$ \\
180 & 1 & 0.2 & 0.1 & 5 & 0.0260 & $0.2761 \pm 0.1271$ & 0.0317 & 0.3492 & 9.43 & -24.53 & $0.4755 \pm 0.1096$ \\
180 & 2 & 0.05 & 0.1 & 5 & 0.0345 & $0.2990 \pm 0.0924$ & 0.0448 & 0.4419 & 9.03 & -31.68 & $0.4355 \pm 0.0038$ \\
200 & 0.05 & 1 & 0.1 & 5 & 0.0190 & $0.2019 \pm 0.0953$ & 0.0234 & 0.2312 & 9.13 & -27.02 & $0.4610 \pm 0.0519$ \\
240 & 0.04 & 1 & 0.1 & 5 & 0.0008 & $0.0135 \pm 0.0614$ & -0.0027 & -0.0319 & 4.01 & -33.71 & $0.1882 \pm 0.1156$ \\
240 & 1 & 0.05 & 0.1 & 5 & 0.0103 & $0.1194 \pm 0.0423$ & 0.0188 & 0.1974 & 2.83 & -34.55 & $0.1368 \pm 0.0567$ \\
\midrule
\multicolumn{12}{c}{\textbf{S\&P 500}} \\
\midrule
$W$ & $\lambda$ & $\gamma$ & $\tau_q$ & $N$ & IC & ICIR & RankIC & RankICIR & AR(\%) & MDD(\%) & Sharpe \\
\midrule
120 & 0.04 & 1 & 0.1 & 5 & 0.0293 & $0.2288 \pm 0.0153$ & 0.0072 & 0.0581 & 17.13 & -26.07 & $0.7533 \pm 0.0415$ \\
120 & 1 & 0.05 & 0.1 & 5 & 0.0253 & $0.1953 \pm 0.0186$ & 0.0066 & 0.0515 & 18.61 & -22.85 & $0.8034 \pm 0.0653$ \\
180 & 0.02 & 1 & 0.1 & 5 & 0.0147 & $0.1083 \pm 0.1164$ & -0.0039 & -0.0292 & 14.12 & -25.13 & $0.6163 \pm 0.2397$ \\
180 & 0.04 & 0.05 & 0.1 & 5 & 0.0226 & $0.1818 \pm 0.0029$ & 0.0033 & 0.0381 & 16.14 & -23.80 & $0.7385 \pm 0.0703$ \\
180 & 0.04 & 1 & 0.02 & 5 & 0.0211 & $0.1661 \pm 0.0020$ & 0.0061 & 0.0450 & 15.53 & -25.82 & $0.6927 \pm 0.0863$ \\
180 & 0.04 & 1 & 0.02 & 10 & 0.0201 & $0.1574 \pm 0.0282$ & 0.0018 & 0.0134 & 16.36 & -25.18 & $0.7156 \pm 0.0066$ \\
180 & 0.04 & 1 & 0.1 & 5 & 0.0246 & $0.1866 \pm 0.0681$ & 0.0052 & 0.0394 & 18.29 & -25.33 & $0.8049 \pm 0.0954$ \\
180 & 0.08 & 1 & 0.1 & 5 & 0.0158 & $0.1252 \pm 0.0332$ & 0.0005 & 0.0039 & 15.35 & -25.06 & $0.6803 \pm 0.0047$ \\
180 & 0.4 & 0.05 & 0.1 & 5 & 0.0296 & $0.2136 \pm 0.0045$ & 0.0075 & 0.0556 & 18.16 & -26.87 & $0.7528 \pm 0.0326$ \\
180 & 1 & 0.05 & 0.1 & 5 & 0.0288 & $0.2067 \pm 0.0102$ & 0.0033 & 0.0252 & 18.90 & -25.67 & $0.7977 \pm 0.0629$ \\
180 & 1 & 0.2 & 0.1 & 5 & 0.0236 & $0.1743 \pm 0.0234$ & 0.0007 & 0.0047 & 17.38 & -25.45 & $0.7494 \pm 0.0316$ \\
180 & 2 & 0.05 & 0.1 & 5 & 0.0216 & $0.1656 \pm 0.0839$ & 0.0019 & 0.0148 & 16.14 & -23.73 & $0.7310 \pm 0.2447$ \\
200 & 0.05 & 1 & 0.1 & 5 & 0.0257 & $0.1928 \pm 0.0142$ & 0.0050 & 0.0382 & 18.82 & -26.68 & $0.8005 \pm 0.0212$ \\
240 & 0.04 & 1 & 0.1 & 5 & 0.0276 & $0.2104 \pm 0.0055$ & 0.0071 & 0.0551 & 17.30 & -24.58 & $0.7533 \pm 0.0369$ \\
240 & 1 & 0.05 & 0.1 & 5 & 0.0234 & $0.1792 \pm 0.0000$ & 0.0060 & 0.0440 & 16.42 & -24.64 & $0.7198 \pm 0.0165$ \\
\bottomrule
\end{tabular}
\end{table*}

\paragraph{Timing and strength of memory reuse.}
Warmup and residual weight control when experience begins to influence search and how strongly it changes edit preferences.
The comparisons in this paragraph fix the quality threshold at 0.10 and batch size at five.
At warmup 180 and failure weight 1.0, increasing the residual weight from 0.02 to 0.08 raises mean ICIR on CSI 500, while S\&P 500 favors the intermediate weight 0.04.
At residual weight 0.04 and failure weight 1.0, the longest warmup reduces CSI 500 ICIR relative to the shorter warmups, whereas S\&P 500 remains competitive at both 120 and 240 attempts.
These patterns suggest that collecting more experience before reuse and applying stronger corrections are not uniformly beneficial.
Memory must balance the reliability of accumulated evidence against the remaining budget for guided exploration.

\paragraph{Failure suppression and admission.}
With warmup 180, residual weight 1.0, quality threshold 0.10, and batch size five, increasing the failure weight from 0.05 to 0.20 improves mean ICIR and Sharpe on CSI 500, but lowers both on S\&P 500.
Thus the strength of failure suppression affects the balance between avoiding repeated rejection and retaining potentially useful edits.
At residual weight 0.04 and failure weight 1.0, lowering the quality threshold improves CSI 500 mean ICIR slightly but reduces it on S\&P 500.
Under this lower threshold, increasing batch size from five to ten reduces mean ICIR on both markets.
Together, these comparisons favor calibrating memory reuse alongside admission and batching, rather than assuming that a more permissive search always produces better factors.

\paragraph{Predictive and portfolio trade-offs.}
The highest mean ICIR and highest mean Sharpe occur at different settings within each market.
This is consistent with the metrics measuring different outcomes: cross-sectional association over all assets and realized returns from a selected portfolio.
The sweep therefore complements the matched ablations by describing how memory controls affect the resulting factor collections.
Rows that change several parameters characterize joint settings, while the conditional comparisons above hold the remaining parameters fixed.

\subsection{Baseline Evaluation Details}

\paragraph{Repetition scope.}
The market-specific AlphaMemo entry on CSI 500 averages 3 complete search runs. The market-specific AlphaMemo entry on S\&P 500 averages 3 complete search runs. AlphaSAGE reports means over three independent search runs per market. Matched AlphaMemo controls use three confirmation seeds, disjoint from scheduler-selection seeds. For mining methods, $n$ counts completed independent search runs. For predictive baselines, $n$ counts predictor fits with fixed inputs and evaluation settings. Alpha158, LightGBM, and LSTM each report $n=3$ predictor fits per market. GP and AlphaGPT each evaluate one saved candidate bank per market. AlphaGen uses 3 completed search runs per market. AlphaAgent uses 2 completed search runs per market. Search-run variation and predictor-fit variation are distinct sources of uncertainty.

\paragraph{Market-specific settings.}
The main AlphaMemo entries use warmup 180, residual weight 2.0, and failure weight 0.05 on CSI 500, and warmup 240, residual weight 0.04, and failure weight 1.0 on S\&P 500. Both use 500 search attempts, batch size five, quality threshold 0.10, and the same labels, purging, factor selection, predictor, and portfolio settings.
Table~\ref{tab:scheduler_sensitivity} reports the complete configuration sweep. The main CSI 500 entry averages 3 complete search runs, and the S\&P 500 entry averages 3. All completed prespecified repeats are included. Matched component ablations use the shared reference scheduler with warmup 180, residual weight 0.04, and failure weight 1.0 on three confirmation seeds.

\paragraph{Shared downstream evaluation.}
The comparison uses the 20-trading-day forward label, common market splits, and the same downstream TopK portfolio evaluator.
For SSPM candidate banks, downstream factor selection removes duplicate expressions, ranks valid factors by search-period $|\mathrm{ICIR}|$, requires at least 80 valid metric days, and retains at most 50 factors.
Mining policies and operator semantics remain method specific. A common evaluation interface does not imply that external systems share the AlphaMemo search loop or generation budget.

\paragraph{Predictive-model inputs.}
Alpha158 is the Qlib expert feature collection evaluated with a LightGBM predictor.
The LightGBM and LSTM baseline interfaces instead use four compact price--volume features: intraday return, daily close return, volume relative to its 20-day mean, and normalized daily range.
Thus they share an input set with each other, not the 158-dimensional Alpha158 feature set.
Predictive fits use the purged splits, with repetition counts given above. LSTM uses 80 training epochs.

\paragraph{Saved candidate banks.}
GP and AlphaGPT use their previously generated market-specific candidate banks, with 500 and 300 attempts, respectively, from the 2016--2020 search period.
Their factors are selected again and their downstream predictors are refitted with the same purged splits and portfolio settings as AlphaMemo.
The saved banks are held fixed for common-protocol evaluation, with mining attempts distinguished from downstream predictor refits.

\paragraph{AlphaSAGE reproduction.}
AlphaSAGE uses its released GFlowNet implementation with 10,000 training episodes and a pool capacity of 50, separately for each market and seeds 0, 1, and 2.
Native expressions are evaluated with the released operators before export, rather than translated into the AlphaMemo parser.
Invalid or nonfinite factors are recorded in the export diagnostics.
The resulting factor collections use the same downstream predictor configuration and label-boundary purging as AlphaMemo.
The shared-predictor evaluation isolates the predictive utility of AlphaSAGE's native factor collections.

\paragraph{AlphaGen interface.}
The released PPO implementation uses its LSTM policy encoder, native symbolic operators, MSE-based alpha pool, and no LLM augmentation.
The run budget is 300,000 environment steps with pool capacity 50.
Training uses the purged 2016--2020 signal dates and the same next-close-aligned 20-day label as the downstream evaluator.
The final native factor collection uses the common predictor for downstream score generation and portfolio construction.

\paragraph{AlphaAgent interface.}
The released hypothesis generator, expression regularizer, and LLM feedback generator are retained, while factor execution and evaluation use a common-protocol adapter.
The adapter evaluates the released function library without translating its expressions into the AlphaMemo operator set.
Nonpositive time windows are rejected to prevent forward price references.
It uses 50 evaluated hypothesis rounds and retains the released method-specific number of expressions per round.
The execution interface normalizes enclosing JSON fences and casts Boolean inputs to numeric values for the released sign operator.
The factor-response example uses valid JSON with the released fields, and construction recovery passes the native regularizer's recorded rejection feedback to the generator without changing its thresholds.
Response and construction recovery allows up to 24 completion calls per attempt and six attempts per evaluated round, with a response limit of 8192 tokens. Attempt-level failures are logged separately and excluded from the evaluated-round count.
During mining, predictor fitting uses 2016--2019 and feedback uses 2020, both within the shared search period.
After mining, distinct valid proposals are ranked on 2016--2020 and evaluated through the common downstream predictor.
This interface compares AlphaAgent's native factor collections using the shared downstream prediction and portfolio protocol.

\paragraph{LLM-based baseline interfaces.}
AlphaGPT's generated formulas are evaluated with a downstream adapter that reports the shared predictive and portfolio metric suite.
The saved AlphaGPT generation runs use the same \texttt{deepseek/deepseek-v4-flash} backend as AlphaMemo.
Their prompts, generation loops, and candidate counts are method specific, so memory attribution relies on the matched internal controls rather than these external comparisons.

\paragraph{Artifact terms and intended use.}
We use released datasets, libraries, and baseline implementations for academic benchmarking, cite their creators, and do not redistribute third-party datasets or baseline code.
The code release is intended for research reproducibility, not financial advice or live trading.

\subsection{Search and Evaluation Settings}

\paragraph{Search and export budgets.}
Each AlphaMemo search makes 500 attempts in batches of five.
The reference scheduler uses a 180-attempt warmup, residual weight 0.04, failure weight 1.0, four sampled motifs, and random motif probability 0.35.
Downstream evaluation ranks distinct valid candidates by search-period $|\mathrm{ICIR}|$ and retains at most 50 factors with at least 80 valid metric days.
Matched controls share the generator, parent scorer, search budget, admission rule, and downstream evaluator.
Invalid or duplicate proposals consume the budget but do not update numerical credit or admission-failure statistics.

\paragraph{Search admission and credit.}
The search-period feedback is $Q=|\mathrm{ICIR}|$ from 2016--2020.
Admission requires $Q\geq0.10$ and maximum absolute Pearson correlation no greater than 0.70 with the admitted pool.
Numerical credit uses distinct, valid parent--child transitions, including children that fail admission.
Invalid and duplicate attempts do not count as numerical failures, and credit is updated under the parent's category with the realized structural motif.
The no-memory control keeps the same bookkeeping but removes both credit corrections and failure penalties from edit scheduling.

\paragraph{Forward-label boundaries.}
The label runs from the next close to the close 20 trading days later.
The last 21 signal dates of each train, validation, and evaluation segment are removed so the label endpoint stays within that segment.
The same purging is used when scoring candidate factors for downstream selection.
The online memory receives only search-period feedback, never downstream validation or evaluation labels.

\paragraph{Shared scheduler selection.}
The reference scheduler is selected using signed downstream validation ICIR, averaged over both markets and selection seeds 0 and 1.
Confirmation seeds 2, 3, and 4 are used for every matched control.
A five-configuration sensitivity check varies warmup over 120, 180, and 240 attempts and residual weight over 0.02, 0.04, and 0.08, holding the failure penalty fixed.
Under the same equal-market validation criterion, the selected 180-attempt, 0.04-weight scheduler scores 0.1757, compared with 0.1476 for the next-best candidate.
The shared reference configuration is retained for the matched controls.
A follow-up validation sweep holds warmup and memory weight fixed while varying admission threshold and batch size.
The equal-market validation scores are 0.1757 for threshold 0.10 with batch size five, 0.0685 for threshold 0.02 with batch size five, and 0.0052 for threshold 0.02 with batch size ten.
The threshold and batch size are retained for both markets.

\begin{table}[t]
\centering
\small
\caption{Shared AlphaMemo configuration used in the matched ablations.}
\label{tab:appendix_operating_points}
\begin{tabular}{lc}
\toprule
\textbf{Parameter} & \textbf{Value} \\
\midrule
Attempts / batch size & 500 / 5 \\
Warmup attempts & 180 \\
Residual-memory weight & 0.04 \\
Failure-memory weight & 1.0 \\
Sampled motifs & 4 \\
Random motif probability & 0.35 \\
Proposal-noise standard deviation & 0.01 \\
Quality / correlation admission & 0.10 / 0.70 \\
Maximum downstream factors & 50 \\
Minimum valid metric days & 80 \\
Selection seeds & 0, 1 \\
Confirmation seeds & 2, 3, 4 \\
\bottomrule
\end{tabular}
\end{table}

\paragraph{Compute environment.}
Experiments use an Intel Core i9-13900 workstation with 64GB RAM and one 24GB NVIDIA RTX 4090.
Formula evaluation, Qlib backtesting, and the reported predictive-model fits use the CPU. LLM generation uses a remote API.

\paragraph{Generator and completion handling.}
AlphaMemo uses \texttt{deepseek/deepseek-v4-flash} through the OpenRouter-compatible interface, temperature 0.7, a maximum output length of 1,024 tokens, and disabled reasoning.
The prompt gives the parent formula, target category, requested motif, supported features and operators, and a single-formula output constraint.
Malformed completions receive bounded retries. Unrecoverable formatting failures consume one attempt without inserting a heuristic substitute or updating numerical credit.
Sustained failure triggers an explicit stop, and partial searches retain replayable event and completion records.

\section{Representative Evolved Factors}
\label{app:factor_examples}

Table~\ref{tab:factor_examples_app} presents representative AlphaMemo factors selected on CSI 500 in a separate search run.
These examples illustrate formula structure. Component effects are evaluated through the matched credit-assignment experiment.
We report absolute predictive metrics because factors with negative IC can be sign-flipped before entering a factor pool.

\begin{table*}[t]
\centering
\scriptsize
\caption{Representative AlphaMemo factors on CSI 500. Metrics are computed on the search split.}
\label{tab:factor_examples_app}
\setlength{\tabcolsep}{3pt}
\renewcommand{\arraystretch}{1.12}
\begin{tabular}{l p{0.48\textwidth} p{0.18\textwidth} c c c}
\toprule
Factor & Formula & Interpretation & $|\mathrm{IC}|$ & $|\mathrm{ICIR}|$ & $|\mathrm{RankICIR}|$ \\
\midrule
SSPM\_000 & \path|CsRank(TsMin(Div(TsSum(Where(Greater($close,Delay($close,5)),Delta(Log(Add($volume,1.0)),5),0.0),10),TsStd(TsSum(Where(Greater($close,Delay($close,5)),Delta(Log(Add($volume,1.0)),5),0.0),10),60)),20))| & temporal price change with cross-sectional/temporal ranking & 0.0325 & 0.4461 & 0.5237 \\
SSPM\_017 & \path|TsRank(Where(Greater($close,Delay($close,5)),TsMean(Log(Add($volume,1.0)),5),0.0),20)| & temporal price change with temporal volume level & 0.0268 & 0.2877 & 0.1979 \\
SSPM\_033 & \path|CsRank(Add(Where(Greater($close,Delay($close,5)),Delta(Log(Add($volume,1.0)),5),0.0),Where(Less($close,Delay($close,5)),Neg(Delta(Log(Add($volume,1.0)),5)),0.0)))| & signed volume response to price direction & 0.0159 & 0.2310 & 0.3745 \\
SSPM\_038 & \path|CsRank(Add(CsRank(Where(Greater($close,Delay($close,5)),Delta(Log(Add($volume,1.0)),5),Neg(Delta(Log(Add($volume,1.0)),5)))),CsRank(Mul(Delta(Log(Add($volume,1.0)),5),Delta($close,5)))))| & rank aggregation with price-volume interaction & 0.0173 & 0.2243 & 0.3743 \\
SSPM\_036 & \path|CsRank(Add(CsRank(Where(Greater($close,Delay($close,10)),Delta(Log(Add($volume,1.0)),10),0.0)),CsRank(Mul(Delta(Log(Add($volume,1.0)),5),Delta($close,5)))))| & longer-horizon price condition and volume-price interaction & 0.0191 & 0.2299 & 0.3881 \\
\bottomrule
\end{tabular}
\end{table*}

\section{More Discussions}
\label{sec:more_discussion}

\noindent$\triangleright$ \textbf{\textit{Q1. Why is process-level memory more suitable than final-factor memory for self-evolving alpha mining?}}

Final-factor memory retrieves what has worked before, but self-evolution needs to know how useful factors were reached.
In alpha mining, two factors with similar search-period scores may arise from very different edits, and a historically strong expression may decay after a market regime shift.
AlphaMemo therefore stores reusable evidence about factor-category contexts, realized structural edit motifs, and edit outcomes.
This moves the persistent state closer to the agent's action space: instead of memorizing only winners, the agent remembers which search moves tend to help or fail.

\noindent$\triangleright$ \textbf{\textit{Q2. How does AlphaMemo limit reliance on uncertain memory?}}

AlphaMemo is designed so that memory is a calibrated residual correction, not an unconditional policy.
The base search ledger selects parents from quality, depth, and retrieval frequency, while admission separately enforces diversity.
SSPM guides the subsequent edit choice rather than reranking all parent-edit pairs. Confidence gates discount low-count or high-variance credit estimates. Failure memory penalizes a category-motif pair after at least five observations and suppresses its credit correction when the empirical failure rate reaches 0.8.
This asymmetry favors avoiding repeatedly observed mistakes over aggressively exploiting early successes.

\noindent$\triangleright$ \textbf{\textit{Q3. Is AlphaMemo tied to a specific generator, market, or evaluator?}}

The memory is attached to the symbolic search process rather than to a particular LLM response format.
It observes executable formulas, parent-child edits, search-period feedback, and admission decisions.
Therefore, the same mechanism can be used with different LLM backbones, heuristic generators, or external formula-generation systems as long as their outputs can be parsed and evaluated.
The CSI 500 and S\&P 500 experiments provide a compact cross-market test of this design.

\noindent$\triangleright$ \textbf{\textit{Q4. Are the gains simply caused by a stronger search ledger rather than memory?}}

The matched component study is constructed to separate these effects.
AlphaMemo and the no-memory control share the generator interface, parent scorer, operator set, downstream factor-selection cap, diversity threshold, and admission rule.
The no-memory control removes both credit corrections and failure penalties. Absolute credit changes the credit target, no-confidence removes confidence attenuation, and no-veto removes admission-failure suppression.
Thus, the mechanism evidence comes from changing the memory component while holding the underlying search machinery fixed.

\noindent$\triangleright$ \textbf{\textit{Q5. Does AlphaMemo require delicate hyperparameter tuning?}}

Scheduler sensitivity shows that the strength and timing of memory influence the results.
The matched ablations use one shared configuration, while the main comparison uses market-specific settings. Appendix~\ref{app:more_exp} lists both.
The matched controls reveal cross-market variation: residual credit improves mean CSI 500 ICIR relative to no memory, but the same predictive improvement does not hold on S\&P 500.
Confidence attenuation and failure suppression are consequently controls on experience reuse, not guarantees against overfitting or adverse calibration.

\noindent$\triangleright$ \textbf{\textit{Q6. Why not use policy fine-tuning, reinforcement learning, or full trajectory replay instead?}}

Policy fine-tuning and reinforcement learning offer complementary ways to improve search, but require a separate policy-training loop.
Trajectory-based memory retains richer end-to-end history, whereas AlphaMemo focuses on reusable editing experience.
It compresses search history into structured category-edit-outcome statistics and applies them as residual evidence at inference time.
New observations update the memory statistics while the generator weights remain fixed.

\noindent$\triangleright$ \textbf{\textit{Q7. How can AlphaMemo complement existing alpha-mining systems?}}

AlphaMemo is most naturally viewed as a memory layer for iterative formula search.
It can sit on top of human-designed factor libraries, symbolic search, or LLM-based generators by recording which edits are productive under repeated evaluation.
This is especially useful when the generator is strong but feedback is sparse and noisy: the memory layer filters repeated failures, encourages reusable motifs, and leaves the generator free to propose novel expressions.

\noindent$\triangleright$ \textbf{\textit{Q8. What remains outside the current scope?}}

AlphaMemo still depends on a reliable evaluator, a well-defined symbolic operator grammar, and enough repeated edit observations to estimate memory statistics.
Repeated observations are needed to distinguish reusable editing experience from noisy individual outcomes.
When observations are sparse, confidence attenuation limits the influence of uncertain residual estimates.
Future work can examine richer market states, multi-horizon objectives, and live paper-trading evaluation.

\section{Statements on AI Assistants in Research and Writing}
\label{app:ai-usage}

We used LLM assistants for language polishing and code refactoring during this work. All scientific claims, experimental results, and analyses were produced and verified by the authors. No AI-generated text was included without author review.

\end{document}